\documentclass[11pt]{article}

\usepackage[final]{acl}

\usepackage{times}
\usepackage{latexsym}

\usepackage[T1]{fontenc}
\usepackage{amsmath}
\usepackage[utf8]{inputenc}

\usepackage{microtype}
\usepackage{siunitx}
\usepackage{inconsolata}

\usepackage{graphicx}
\usepackage{booktabs}
\usepackage{multirow}
\usepackage{adjustbox}
\usepackage{wrapfig}
\usepackage{dsfont}
\usepackage{placeins}
\usepackage{amssymb}
\usepackage[normalem]{ulem}
\usepackage[table]{xcolor}
\definecolor{lightgreen}{RGB}{171,208,241}

\usepackage[most]{tcolorbox} 
\usepackage{listings}      

\newtcolorbox{promptbox}[1]{
    colback=white,                % 框内主体的背景色（白色）
    colframe=gray!80!black,       % 边框颜色（它会自动作为标题栏的背景色，深灰）
    coltitle=white,               % 标题文字的颜色（白色）
    fonttitle=\sffamily\bfseries, % 标题字体（无衬线加粗）
    title={#1},                   % 标题内容
    boxrule=0.5pt,                % 边框的粗细
    arc=2mm,                      % 整体外边框的圆角弧度
    left=2mm, right=2mm,          % 正文左右留白
    top=2mm, bottom=2mm,          % 正文上下留白
    fontupper=\small\ttfamily     % 正文字体（小号、等宽）
}

\title{{R2IF: Aligning Reasoning with Decisions via Composite Rewards for Interpretable LLM Function Calling}}

\author{
    Aijia Cheng\textsuperscript{1},
    Kailong Wang\textsuperscript{2},
    Ling Shi\textsuperscript{3},
    Yongxin Zhao\textsuperscript{1}\thanks{Corresponding author.} \\
    \textsuperscript{1}Shanghai Key Laboratory of Trustworthy Computing, \\
    East China Normal University, Shanghai, China \\
    \textsuperscript{2}Huazhong University of Science and Technology \\
    \textsuperscript{3}Nanyang Technological University \\
    \texttt{51275902045@stu.ecnu.edu.cn},
    \texttt{wangkl@hust.edu.cn},\\
    \texttt{ling.shi@ntu.edu.sg},
    \texttt{yxzhao@sei.ecnu.edu.cn}
}

\begin{document}
\maketitle
\begin{abstract}
Function calling empowers large language models (LLMs) to interface with external tools, yet existing RL-based approaches suffer from misalignment between reasoning processes and tool-call decisions. We propose R2IF, a reasoning-aware RL framework for interpretable function calling, adopting a composite reward integrating format/correctness constraints, Chain-of-Thought Effectiveness Reward (CER), and Specification-Modification-Value (SMV) reward, optimized via GRPO. Experiments on BFCL/ACEBench show R2IF outperforms baselines by up to 34.62\% (Llama3.2-3B on BFCL) with positive Average CoT Effectiveness (0.05 for Llama3.2-3B), enhancing both function-calling accuracy and interpretability for reliable tool-augmented LLM deployment.
\end{abstract}

\section{Introduction}

Large language models (LLMs) excel at open-domain text generation but lack reliability in real-time structured interactions (e.g., tool invocation) due to static offline training data~\cite{liu2024apigen,liu2024toolace,wang2024hammerbench}. Function calling addresses this by translating natural language queries into precise executable tool APIs, unlocking up-to-date knowledge and complex task-solving capabilities.

% Although large language models (LLMs) have performed strong performance in text generation task, their knowledge which is learned from offline training data leads to unreliable response in the scenario where real-time information is necessary. Function calling addresses this difficulty by converting natural query into executable interfaces, enabling LLMs to establish reliable interaction with external systems and acquire the latest knowledge. This capability allows LLMs to adapt to complex and diverse external environments and complete a greater variety of tasks.  

\begin{figure}[t]
  \centering \includegraphics[width=0.8\columnwidth]{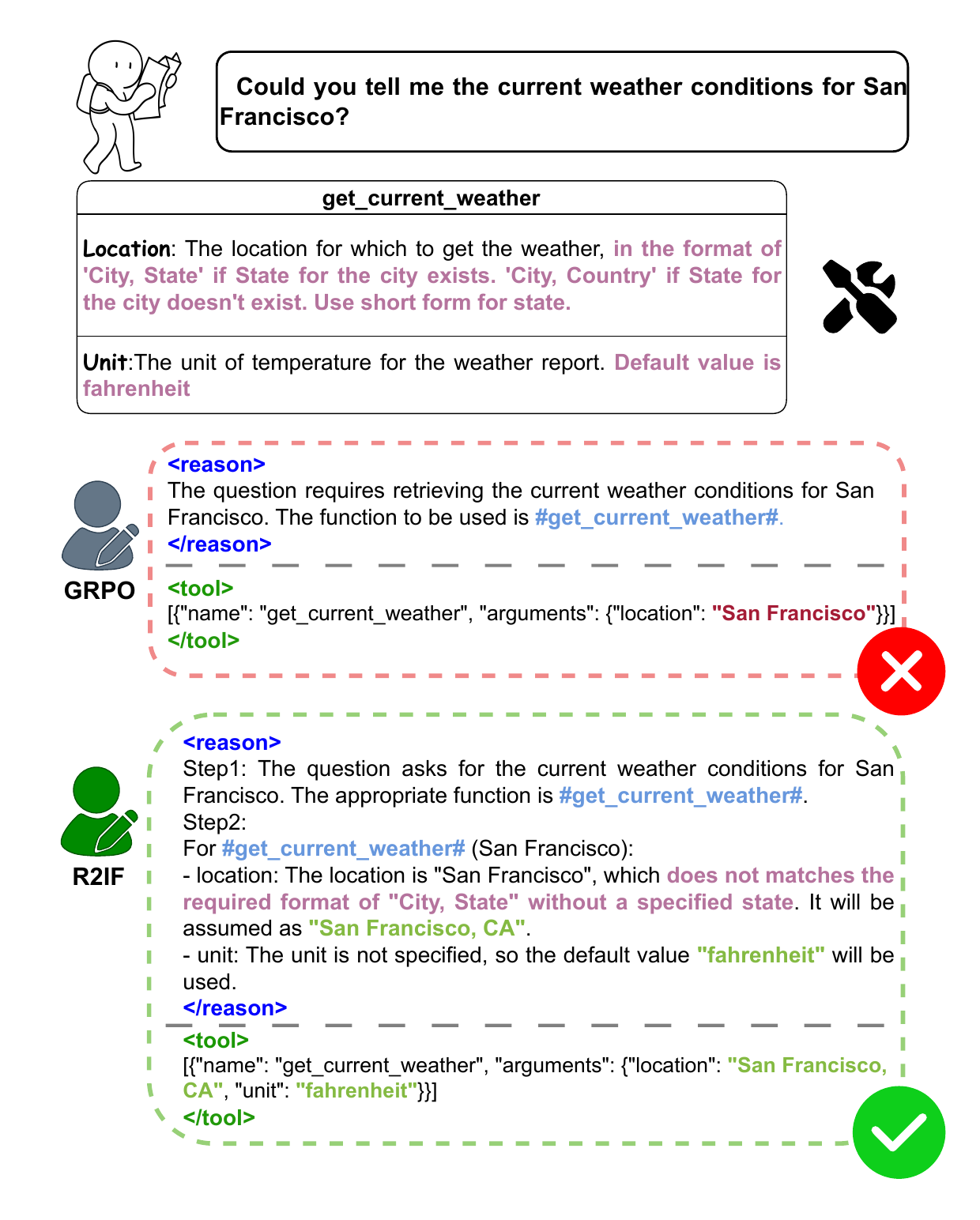}
  \caption{Our method correctly supplements parameter values (e.g., CA, fahrenheit) by aligning reasoning with tool specifications, while the baseline GRPO model fails to address parameter format requirements.}
  \label{fig:example}
\end{figure}

Early function calling research relied on supervised fine-tuning (SFT)~\cite{hao2025funreason} with annotated tool-use data, but was bottlenecked by scarce high-quality executable data. As training signals and evaluation protocols matured, reinforcement learning (RL)~\cite{qian2025toolrl,zhang2025tooln1} emerged as a powerful alternative, directly optimizing functional correctness and enabling accurate function calls across diverse benchmarks. Yet most RL-based methods remain outcome-driven, rewarding only final correctness via AST evaluation—correct calls~\cite{patil2025bfcl} do not equate to sound reasoning, as generated reasoning often serves as post-hoc rationalization rather than guiding tool selection or parameter construction. For example, when queried about San Francisco's weather (Figure \ref{fig:example}), the baseline GRPO model's reasoning only identifies \#get\_current\_weather\# but ignores parameter format (City, State) and default values, leading to incomplete tool calls. Given function calling's structural rigor and sensitivity to fine-grained arguments, accuracy alone fails to capture reasoning quality or system robustness.

% \sout{execution accuracy in function calling. Under this paradigm, modern LLMs are able to generate function calls with correct function names and arguments across a wide range of benchmarks, achieving strong overall performance.}

% Despite these advances, most RL-based approaches to function calling remain outcome-driven, rewarding only final result correctness with AST Evaluation. This implicitly assumes that correct calls reflect correct reasoning, yet in practice the generated reasoning often becomes post-hoc rationalization rather than a driver of tool selection or parameter construction. Since function calling is highly structured and sensitive to fine-grained argument specifications, accuracy alone provides an incomplete proxy for reasoning quality and system reliability.

% A natural remedy is to reward the reasoning process directly, but this is non-trivial. Without explicit grounding in tool schemas and parameter transformations, process rewards can collapse into generic textual signals that incentivize verbosity over decision-supportive reasoning. Worse, unconstrained reasoning rewards may degrade parameter-level precision by reinforcing plausible-looking but action-misaligned rationales, so “rewarding reasoning” does not necessarily yield more reliable tool-use behavior.

Directly rewarding reasoning without grounding in tool schemas is non-trivial. Unlinked to tool specifications and parameter transformations, process rewards prioritize verbosity over utility. Worse, unconstrained reasoning rewards may erode parameter precision by reinforcing plausible but tool-call decision-misaligned rationales, making "rewarding reasoning" insufficient for reliable tool use.

% These issues expose a key gap: we lack a reward framework that preserves execution correctness while making reasoning substantively influence function-calling decisions. Effective supervision should differentiate reasoning that merely accompanies a correct action from reasoning that supports specification, modification, and value instantiation. Addressing this gap is crucial for both interpretability and the robustness of tool-augmented LLMs in real-world deployment.

Addressing this gap is pivotal to enhancing tool-augmented LLMs' interpretability and robustness: ungrounded reasoning complicates error diagnosis, causes failures in unseen parameter formats or queries, and undermines trust and generalization. Effective supervision must distinguish between reasoning that merely accompanies correct actions and reasoning that actively supports parameter specification, necessary modifications, and value instantiation—three pillars of reliable tool use.

% In this work, we propose R2IF, a reasoning-aware reinforcement learning framework for interpretable function calling. We optimize a composite reward that combines (i) a binary reward for strict format and exact-match tool-call correctness, (ii) a CoT effectiveness reward that measures whether the reasoning prefix improves tool-call success, and (iii) an SMV reward that aligns reasoning with parameter specifications, required modifications, and value instantiation. By treating function calling as a sequence of parameter-level decisions, R2IF provides supervision beyond final outcomes and discourages correct-but-ungrounded rationales. Experiments show that R2IF consistently improves function-calling accuracy and yields reasoning that is more stable and better aligned with executable decisions.

\noindent\textbf{Our Work.} In this work, we propose R2IF, a reasoning-aware reinforcement learning framework for interpretable function calling in large language models. We formalize interpretability as the alignment between reasoning and executable tool-call decisions, grounded in parameter specification, modification, and value instantiation. Building on this, we design a composite reward integrating strict format/correctness constraints, a distribution-based Chain-of-Thought~\cite{wei2023chainofthoughtpromptingelicitsreasoning} Effectiveness Reward (CER) quantifying reasoning utility, and a Specification-Modification-Value (SMV) reward enforcing parameter-level reasoning alignment. Optimized via Grouped Proximal Policy Optimization (GRPO)~\cite{shao2024deepseekmath}, R2IF provides trajectory-level supervision to discourage correct-but-ungrounded tool calls and strengthen reasoning-tool call decision links.

\noindent\textbf{Results and Findings.} Across BFCL and ACEBench benchmarks, R2IF achieves top overall accuracy across Qwen2.5 (1.5B/3B/7B) and Llama3.2-3B backbones, with a +34.62\% gain on Llama3.2-3B (BFCL) and +15.4\% on Qwen2.5-7B (ACEBench). It maintains strong irrelevant tool-rejection performance ($\ge$96\% accuracy) while boosting Atomic and Single-turn task gains, confirming balanced improvement over conservative strategies. R2IF's reasoning effectiveness (ACE) turns positive for 3B/7B models (e.g., 0.05 for Llama3.2-3B on BFCL), outperforming baselines with negative values. Notably, R2IF scales monotonically with model size and generalizes across model families, remaining competitive under real-user Live scenario distribution shifts.

\noindent\textbf{Summary of contributions:}
Our contributions are summarized as follows:
% \begin{itemize}
%     \item We propose a hybrid reward design that jointly optimizes the reasoning process and the final function-call result.
%     \item We introduce a distribution-based effectiveness metric to measure whether reasoning is truly helpful and sufficiently informative for the model's decisions.

% \end{itemize}
 \begin{itemize}
\item We propose a hybrid reward design that jointly optimizes reasoning quality and final function-call correctness, addressing the misalignment between reasoning processes and tool-call decisions in RL-based function calling.
\item We introduce a distribution-based Chain-of-Thought Effectiveness Reward (CER), which enhances tool-call stability without relying on subjective reasoning scoring.
\item We design a parameter-level Specification-Modification-Value (SMV) as an optimization of outcome-oriented RL methods, explicitly supervising parameter constraints, transformations, and value instantiation to improve interpretability and precision.
\end{itemize}

\section{Background and Related Work}

\begin{figure*}[t]
  \centering
  \includegraphics[width=0.8\textwidth]{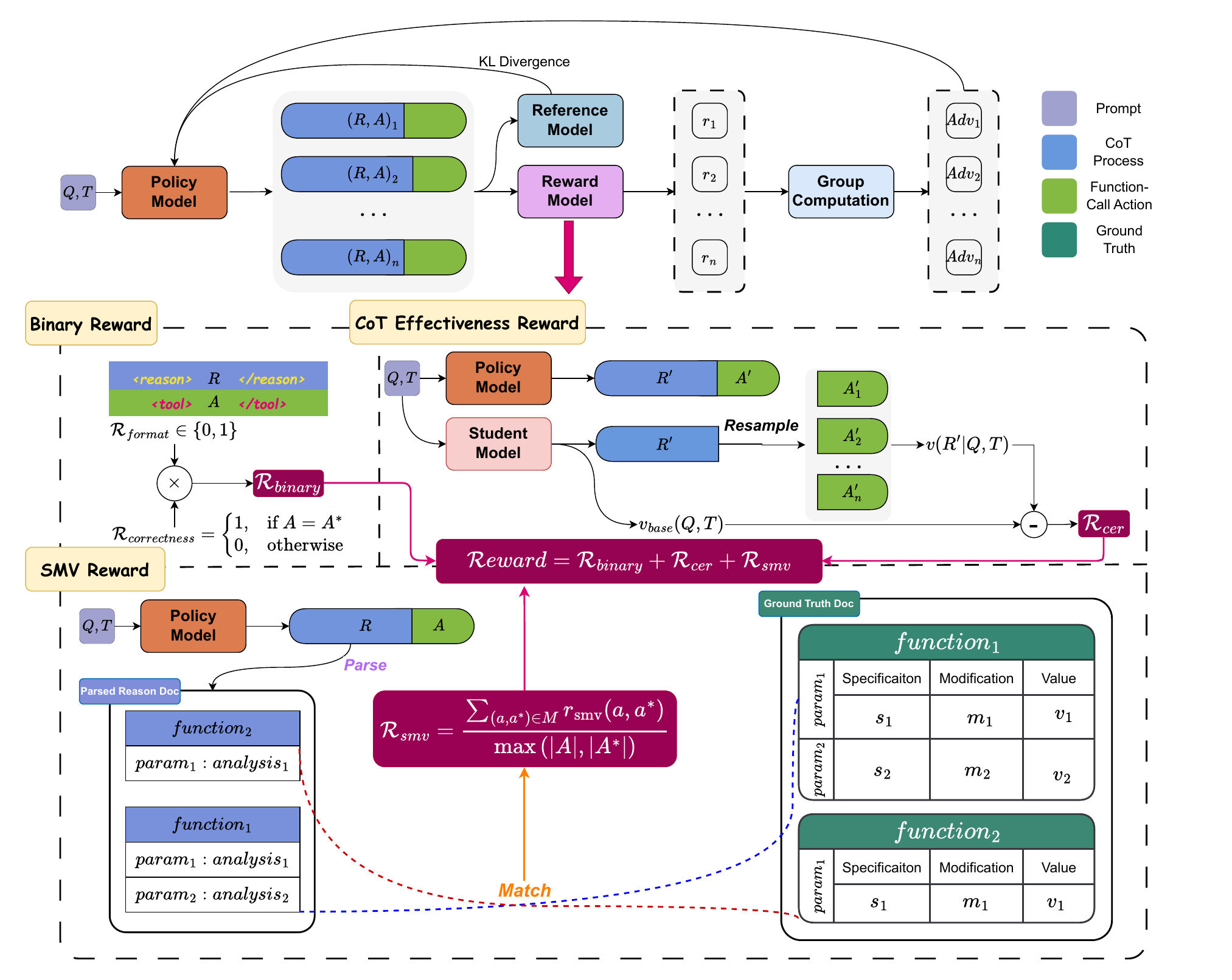}
  \caption{Overall pipeline of our reward optimization. We (i) compute a binary reward to enforce strict function-call correctness, (ii) compute a CER reward to measure whether the CoT supports the chosen actions, and (iii) compute an SMV reward to align the CoT with ground-truth parameter specification, modification, and value.}
  \label{fig:Overview}
\end{figure*}

\paragraph{Formalization.} In the function calling task, our goal is to evaluate the model's ability to deterministically transform user queries into executable actions. The core formalization of this task is as follows: 

Given a user query \( Q \) and a tool set \( T = \{ t_1, t_2, ..., t_n \} \), the model performs a mapping function \( f \), producing an ordered output tuple 
\[(R, A) = f(Q, T). \]

Here, \( R \) represents the reasoning process, which is a text reasoning sequence; \( A \) represents the function call result, which can vary depending on the specific scenario, and it may be an empty action \( \emptyset \), a single action, or a set of actions. The specific scenarios are all presented in Table~\ref{tab:task_categories}.

\begin{table}[t]
\caption{Comparison of task categories in terms of tool availability and invocation patterns.}
\label{tab:task_categories}
\centering

\small
\setlength{\tabcolsep}{6pt}
\renewcommand{\arraystretch}{1.15}

\begin{adjustbox}{max width=\columnwidth}
\begin{tabular}{lccc}
\toprule
\textbf{Task Category} &
\textbf{Tool Set Size $|T|$} &
\textbf{Distinct Tools Used} &
\textbf{\# Tool Calls $|A|$} \\
\midrule
Simple        & $1$          & $1$        & $1$   \\
Multiple           & $>1$         & $1$        & $1$   \\
Parallel           & $1$          & $1$        & $>1$  \\
Parallel Multiple  & $>1$         & $\ge 1$    & $>1$  \\
Irrelevance        & $\mathbb{N}$ & $0$        & $0$   \\
\bottomrule
\end{tabular}
\end{adjustbox}

\end{table}

\paragraph{Function call.} Early work on LLM function calling primarily relied on supervised fine-tuning (SFT). To mitigate the lack of high-quality data, several studies proposed scalable synthetic data pipelines.
\citet{liu2024apigen} introduce APIGen, which generates verifiable function-calling data via hierarchical checks, while~\citet{liu2024toolace} propose ToolACE, an agentic synthesis framework that expands API coverage and scenario complexity.
For standardized evaluation, \citet{patil2025bfcl} present the Berkeley Function Calling Leaderboard (BFCL), which has become a widely adopted benchmark.

More recently, reinforcement learning (RL) has been used to frame function calling as a policy optimization problem, directly optimizing correctness under verifiable rewards. Inspired by R1-style reasoning models~\citep{deepseekai2025r1}, RL has been adapted to tool use, with studies exploring reward design for tool selection and argument construction under GRPO~\citep{qian2025toolrl}, as well as rule-based RL with binary correctness and format rewards~\citep{zhang2025tooln1}.

\paragraph{RL for LLM Reasoning.}
Reinforcement learning has emerged as a core post-training paradigm for enhancing LLM reasoning by optimizing generation policies with outcome-based feedback and improving long-horizon credit assignment~\citep{ouyang2022training}.
In practice, PPO-style optimizers and their variants (e.g., GRPO) are widely used to stabilize and scale reasoning-oriented training~\citep{schulman2017ppo,shao2024deepseekmath,deepseekai2025r1}.
Beyond answer-level rewards, prior work emphasizes process-aware supervision and step-level evaluation to provide denser signals for multi-step reasoning~\citep{lightman2023verify,zheng2025correctnessexposingllmgeneratedlogical}.
Recent studies further extend RL to tool-augmented reasoning, rewarding correct tool usage under structured action spaces~\citep{qian2025toolrl,zhang2025tooln1}.

\paragraph{GRPO.} Each training instance consists of a user query $Q$ and a tool set $T$. 
Given $(Q, T)$, the policy generates multiple rollouts, each producing
\[
(R, A) = f(Q, T),
\]
where $R$ is the reasoning sequence and $A$ is the resulting function call. Collecting $n$ rollouts for the same query yields
\[
G_{Q} = \{((R_1, A_1), r_1), \ldots, ((R_n, A_n), r_n)\},
\]
where $r_i$ is the scalar reward of the $i$-th rollout.

We normalize rewards within each group. The mean and standard deviation are
\[
\mu_{Q} = \frac{1}{n} \sum_{i=1}^n r_i, \quad 
\sigma_{Q} = \sqrt{\frac{1}{n} \sum_{i=1}^n (r_i - \mu_{Q})^2}.
\]
The normalized advantage is defined as
\[
{Adv}_i(R_i, A_i \mid Q) = \frac{r_i - \mu_{Q}}{\sigma_{Q} + \eta},
\]
where $\eta$ is a small constant for numerical stability.

Policy optimization follows the clipped PPO objective:
\[
{\small
\begin{aligned}
J_{\text{GRPO}}(\theta) 
  &= \mathbb{E}_{Q \sim \mathcal{D}} \,
     \mathbb{E}_{(R_i, A_i) \sim \pi_\theta} \Big[
     \min \Big(
       \rho_\theta^{(i)} {Adv}_i(R_i, A_i \mid Q), \\
  &\qquad\operatorname{clip}\big(
       \rho_\theta^{(i)},\,
       1 - \epsilon,\,
       1 + \epsilon
     \big) {Adv}_i(R_i, A_i \mid Q)
     \Big)
     \Big],
\end{aligned}
}
\]
where
\[
\rho_\theta^{(i)} = \frac{\pi_\theta(R_i, A_i \mid Q, T)}{\pi_{\text{old}}(R_i, A_i \mid Q, T)},
\]
and $\epsilon$ is the PPO clipping hyperparameter. Optionally, a KL penalty to a reference policy can be added to further constrain updates.

\section{Method}
In this section, we describe our R2IF  Method in detail, including the design of reward.  An overview of the R2IF framework is presented in Figure \ref{fig:Overview}.

\subsection{Reward Function Design}

To train the policy to produce both executable actions and well-grounded reasoning, we design a composite reward function that combines hard structural constraints with fine-grained reasoning supervision. For a rollout $(R, A)$ generated from a query $Q$ and tool set $T$, the overall reward is defined as
\[
\mathcal{R} = \mathcal{R}_{\text{binary}} + \mathcal{R}_{\text{cer}} + \mathcal{R}_{\text{smv}},
\]
where $\mathcal{R}_{\text{binary}}$ enforces both the format and the correctness of the tool calls, $\mathcal{R}_{\text{cer}}$ is used to reward the overall effectiveness of the reasoning, and $\mathcal{R}_{\text{smv}}$ provides a finer-grained reward signal by evaluating whether the reasoning correctly identifies the parameter specifications, applies the corresponding modifications, and ultimately derives the correct values. Each component is described below.

\subsubsection{Binary Reward}

The binary reward acts as a hard gate that ensures the model output is both well-formatted and exactly consistent with the ground truth. It is defined as
\[
\mathcal{R}_{\text{binary}} =  \mathcal{R}_{\text{format}} \cdot \mathcal{R}_{\text{correctness}},
\]
where $\mathcal{R}_{\text{format}} \in \{0,1\}$ denotes format validity and $\mathcal{R}_{\text{correctness}} \in \{0,1\}$ denotes tool-call correctness. In the final reward aggregation, the binary reward is weighted by 3 to avoid the optimization focus shifting excessively toward the softer reward terms.

\paragraph{Format validity}
$\mathcal{R}_{\text{format}} = 1$ if and only if the output satisfies all of the following conditions:  
(i) it contains exactly one \texttt{<reason></reason>} block and one \texttt{<tool></tool>} block;  
(ii) the \texttt{<reason>} block precedes the \texttt{<tool>} block;
(iii) response should not include any other text. 
Otherwise, $\mathcal{R}_{\text{format}} = 0$.

\paragraph{Tool-call correctness}
Let $A^{*}$ denote the ground-truth action list and $A$ the predicted action list, where each action is represented as a tuple $(\text{name}, \text{arguments})$. We set $$\mathcal{R}_{correctness}=\begin{cases}1,&\text{if }A=A^{*}\\0,&\text{otherwise}\end{cases}$$

For irrelevance-rejection instances, where no tool should be invoked, $\mathcal{R}_{\text{correctness}} = 1$ if the model explicitly outputs the predefined rejection string; otherwise $\mathcal{R}_{\text{correctness}} = 0$.

\subsubsection{Chain-of-Thought Effectiveness Reward}

While $\mathcal{R}_{\text{binary}}$ evaluates the structural correctness of format and tool call result, it does not directly capture whether the generated cot $R$ is effective in supporting correct tool call.
Thus, we introduce the Chain-of-Thought Effectiveness Reward (CER), which estimates the contribution of the reasoning prefix to successful tool call.

For each training instance, we maintain a precomputed success rate as the baseline value $v_{\text{base}}(Q,T)$ of a faithful student model which is finetuned by this faithfulness-enhancing method\cite{akter2025inducingfaithfulnessstructuredreasoning}, representing the student model's own ability. 

Then, for the given a rollout $(R,A)$, we estimate the effectiveness of the reasoning prefix by letting the student  model sample multiple continuations conditioned on the fixed prefix
\[
\texttt{<reason>}R\texttt{</reason><tool>}.
\]
Each sampled continuation produces a candidate action sequence, which is evaluated with the same ground truth. Finally, the success rate over these samples defines the effectiveness of the reasoning process $v(R \mid Q,T)$.

The chain-of-thought effectiveness reward is then defined as the advantage over the baseline:
\begin{equation}
\label{eq:cer}
\mathcal{R}_{\text{cer}} = v(R \mid Q,T) - v_{\text{base}}(Q,T).
\end{equation}

More concretely, $\mathcal{R}_{\text{cer}}$ measures whether the reasoning $R$ can search enough information so that it can not only guide the student model but also itself to generate the correct tool-call result.

\subsubsection{Specification--Modification--Value Reward}
\label{sec:smv}

In function calling, argument values often require non-trivial processing rather than direct extraction from the user query.  Correct tool call therefore depends not only on producing the final argument values, but also on whether the model correctly identifies the underlying constraints and transformations implied by the tool specification.

To capture this structure, we introduce the \textbf{Specification--Modification--Value (SMV) reward}, which evaluates the extent to which the model's reasoning explicitly supports parameter-level decisions. Concretely, SMV measures whether the reasoning (i) correctly identifies the \emph{specification} of an argument, (ii) describes the necessary \emph{modification} from the user query to a valid argument value, and (iii) ultimately materializes the argument \emph{value} in the tool call.

\paragraph{Documents and parsing.}
Given a response with predicted tool calls $A=\{a_i\}_{i=1}^{|A|}$, we parse the chain-of-thought into an \emph{answer reasoning document}
$\mathrm{Doc}^{\mathrm{ans}}=\{\mathrm{doc}^{\mathrm{ans}}_i\}_{i=1}^{|A|}$,
where $\mathrm{doc}^{\mathrm{ans}}_i.\mathrm{arg}[p]$ is the reasoning snippet for argument $p$ in call $a_i$.
Meanwhile we provide a \emph{ground-truth document}
$\mathrm{Doc}^{*}=\{\mathrm{doc}^{*}_j\}_{j=1}^{|A^{*}|}$, where each argument has metadata
$\mathrm{doc}^{*}_j.\mathrm{arg}[p]=\langle s_p, m_p, \cdot\rangle$
(specification $s_p$, modification $m_p$) which is generated by LLM. Details on how the ground-truth document is obtained are provided in Appendix~\ref{prompt4smv}.

\paragraph{Exact-match alignment.}
We compute SMV only on strictly aligned calls.
Let $M=\{(i,j)\}$ be the exact-match set, where $(i,j)\in M$ iff $a_i=a^{*}_j$ where $a^{*}_j \in A^{*}$. 
This prevents SMV credit for mismatched tools or schemas.

\paragraph{Parameter-level SMV.}
For a matched pair $(i,j)\in M$, let
$R_p=\mathrm{doc}^{\mathrm{ans}}_i.\mathrm{arg}[p]$ denote the reasoning snippet for parameter $p$,
and let $\langle s_p,m_p,\cdot\rangle=\mathrm{doc}^{*}_j.\mathrm{arg}[p]$ be the corresponding
ground-truth specification and modification.
We define the parameter-level SMV score as
\begin{equation}
\label{eq:param_smv_short}
r_{\mathrm{smv}}(p)=\tfrac{1}{3}\!\left(
r_{\mathrm{spec}}(p)+r_{\mathrm{mod}}(p)+r_{\mathrm{val}}(p)
\right),
\end{equation}
which equally weights three complementary aspects of parameter-level correctness.

The value term provides a hard executability signal:
\begin{equation}
\label{eq:smv_val_short}
r_{\mathrm{val}}(p)=\mathbb{I}\!\left[p\in a_i.\mathrm{arg}\right],
\end{equation}
indicating whether the parameter $p$ is actually instantiated in the final tool call.

To assess specification and modification awareness, we compute sentence-level semantic similarity
between the reasoning snippet $R_p$ and the ground-truth text $t \in \{s_p, m_p\}$,
with a threshold gate $\tau$:
\begin{equation}
\label{eq:smv_sem_short}
r_{\mathrm{sem}}(p,t)=
\mathbb{I}\!\left[\mathrm{sim}(R_p,t)\ge\tau\right]\cdot \mathrm{sim}(R_p,t).
\end{equation}
We then set $r_{\mathrm{spec}}(p)=r_{\mathrm{sem}}(p,s_p)$ and
$r_{\mathrm{mod}}(p)=r_{\mathrm{sem}}(p,m_p)$ when the corresponding annotations are available,
and $0$ otherwise.
This design rewards reasoning that explicitly acknowledges argument constraints
and required transformations, rather than merely producing the correct final value.

\paragraph{Tool-call-level SMV.}
For $(i,j)\in M$, we average over supervised arguments $P^{*}_j$ in $\mathrm{doc}^{*}_j$:
\begin{equation}
\label{eq:call_smv_short}
r_{\mathrm{smv}}(a_i,a^{*}_j)=
\frac{1}{|P^{*}_j|}
\sum_{p\in P^{*}_j} r_{\mathrm{smv}}(p).
\end{equation}

\paragraph{Action-level SMV.}
We aggregate over matched call pairs and normalize by call-count mismatch (as in the SMV Reward Part of Fig.~\ref{fig:Overview}):
\begin{equation}
\label{eq:smv_overall_short}
\mathcal{R}_{\mathrm{smv}} = 
\frac{\sum_{(i,j)\in M} r_{\mathrm{smv}}(a_i,a^{*}_j)}
{\max(|A|,\ |A^{*}|)}.
\end{equation}

\paragraph{Irrelevance case}
For irrelevance instances where the correct action is an empty tool-call set, we prepare a reason why the tool can not be called and then compare it with the reasoning process $R$ by similarity to check whether the problem is located or not.

\section{Experiment}
\subsection{Training Dataset}
For the training dataset, we select a subset of data from ToolACE dataset\cite{liu2024toolace}. After data filtering and verification, we find that the number of Parallel case is relatively small, with no more than 500 instances. Thus, to avoid the influence of data imbalance, we sample 500 instances from the other four cases and finally get a balanced training set of 2,500 samples in total.

\subsection{Experiment Setting}

\paragraph{Training.}
We use Verl to train models on 6 NVIDIA RTX PRO 6000 GPUs. The details are listed in Table~\ref{tab:grpo_config}.

\paragraph{Baselines.}
We compare our method against the following baselines:
(1) \textbf{Raw Instruct Model} is the original instruction-tuned model.
(2) \textbf{SFT} fine-tunes the raw instruct model using the same training set, with the reasoning process distilled from GPT-4o~\cite{openai2024gpt4ocard}.
(3) \textbf{Binary reward}\cite{zhang2025tooln1} applies GRPO training with only binary reward.
(4) \textbf{ToolRL}\cite{qian2025toolrl} applies GRPO training with the specific designed reward.

\paragraph{Benchmarks.}
We select the single-turn test case from several benchmarks to comprehensively evaluate the tool-calling performance of models.
(1) \textbf{Berkeley Function Calling Leaderboard (BFCL)}~\cite{patil2025bfcl} is a widely adopted benchmark designed to assess large language models' ability to invoke external functions.
(2) \textbf{ACEBench}~\cite{chen2025acebenchwinsmatchpoint} is a more recent comprehensive benchmark for tool usage. 

\paragraph{Metrics.}
We report two metrics to evaluate both tool call correctness and the utility of the generated reasoning.

\textbf{Accuracy} is the standard exact-match metric for tool calling. A prediction is counted as correct if its tool-call output matches the ground-truth under the official evaluation script; otherwise it is incorrect.

\textbf{Average CoT Effectiveness} is a metric to evaluate the overall utility of CoT reasoning process across multiple cases or tasks. It calculates the average effectiveness of the reasoning prefix in assisting tool-call decisions, based on its contribution to improving the model's tool-call performance.

For each test case $i$, we compute the Chain-of-Thought effectiveness using CER in Eq.~\ref{eq:cer}:
\[
R_{\text{cer}}^{(i)} = v(R^{(i)} \mid Q,T) - v_{\text{base}}(Q,T),
\]
where $v(R^{(i)} | Q,T)$ is the student model's success rate conditioned on reasoning prefix $R^{(i)}$, and $v_{\text{base}}(Q,T)$ is its baseline success rate without it.

We report \textbf{Average CoT Effectiveness (ACE)} as the mean CER over all $N$ cases:
\[
\text{ACE} = \frac{1}{N} \sum_{i=1}^{N} R_{\text{cer}}^{(i)}.
\]

\subsection{Main Results}
We report our method and baseline performance on BFCL and ACEBench with Qwen2.5-1.5B-Instruct, Qwen2.5-3B-Instruct, Qwen2.5-7B-Instruct~\cite{team2024qwen2} and Llama3.2-3B~\cite{dubey2024llama} as backbones in  Table~\ref{tab:bfcl-main} and Table~\ref{tab:acebench}.

\paragraph{Result on BFCL.}
From an overall perspective, R2IF demonstrates clear and stable advantages on BFCL, exhibiting strong consistency across both model scales and model families. R2IF achieves the highest Overall accuracy on Qwen2.5-3B, Qwen2.5-7B, and Llama3.2-3B, and attains the second-best Overall result on Qwen2.5-1.5B.

As shown in the table~\ref{tab:bfcl-main}, while maintaining strong performance on the Irrelevance subset, R2IF also brings consistent gains on Non-live settings. This indicates that the improvements enhance the end-to-end function-calling capability rather than a conservative rejection strategy.

Further comparison between the Non-live and Live splits reveals that the Live setting is substantially more challenging, with all models and training strategies suffering performance degradation due to the distribution shift from synthetic data to real user requests. Nevertheless, R2IF remains more competitive on Live-related metrics.

\begin{table}[tb]
\captionsetup{justification=justified,singlelinecheck=false}
\caption{The evaluation results on the BFCL (last updated November 19, 2025). For the best score, we use \textbf{boldface}, and for the second-best score, we use \uline{underline}.}
\label{tab:bfcl-main}
\centering
\footnotesize
\setlength{\tabcolsep}{3pt}

\begin{adjustbox}{center,max width=0.75\columnwidth}
\begin{tabular}{llcccc}
\toprule
\multirow{2}{*}{Model} & \multirow{2}{*}{Method}
& \multicolumn{4}{c}{\textbf{Overall}} \\
\cmidrule(lr){3-6} 

& 
& Non-live & Live & Irrelevance & Overall \\
\midrule

\multirow{5}{*}{Qwen2.5-1.5B-Instruct}
& Raw             & 30.75 & 37.12 & 78.62 & 48.83 \\
& SFT             & \textbf{33.63} & 33.61 & 74.18 & 47.14 \\
& Binary Reward   & \uline{33.25} & \textbf{38.41} & \textbf{98.51} & \textbf{56.72} \\
& ToolRL          & 31.19 & 35.80 & 95.15 & 54.04 \\        

& R2IF     \cellcolor{lightgreen} &\cellcolor{lightgreen} 29.38 &\cellcolor{lightgreen} \uline{37.62} &\cellcolor{lightgreen} \uline{96.17} &\cellcolor{lightgreen} \uline{54.39} \\
\midrule

\multirow{5}{*}{Qwen2.5-3B-Instruct}
& Raw             & 30.50 & 35.09 & 92.44 & 52.68 \\
& SFT             & 31.19 & \uline{40.46} & 83.64 & 51.76 \\
& Binary Reward   & \uline{31.69} & 39.90 & \uline{98.62} & \uline{56.74} \\
& ToolRL          & 28.00 & 27.68 & \textbf{98.64} & 51.44 \\

&\cellcolor{lightgreen} R2IF          & \cellcolor{lightgreen}\textbf{31.81} &\cellcolor{lightgreen} \textbf{43.07} &\cellcolor{lightgreen} 98.25 & \cellcolor{lightgreen}\textbf{57.71} \\
\midrule

\multirow{5}{*}{Qwen2.5-7B-Instruct}
& Raw             & 55.50 & 56.07 & 87.85 & 66.47 \\
& SFT             & 57.13 & 51.25 & \uline{96.00} & 67.96 \\
& Binary Reward   & 55.13 & 59.97 & \textbf{96.85} & 70.65 \\
& ToolRL          & \uline{57.50} & \uline{61.64} & 95.57 & \uline{71.57} \\       

& R2IF      \cellcolor{lightgreen}      & \textbf{59.50}\cellcolor{lightgreen} & \textbf{63.33} \cellcolor{lightgreen}& 93.59 \cellcolor{lightgreen}& \textbf{72.14} \cellcolor{lightgreen} \\
\midrule

\multirow{5}{*}{Llama3.2-3B-Instruct}
& Raw             & 40.50 & 18.19 & 54.08 & 37.59 \\
& SFT             & 60.19 & 33.45 & 75.61 & 56.41 \\
& Binary Reward   & 65.44 & 37.06 & \uline{90.27} & 64.26 \\
& ToolRL          & \uline{68.13} & \uline{45.00} & 85.65 & \uline{66.26} \\        

& R2IF    \cellcolor{lightgreen}      &\cellcolor{lightgreen} \textbf{69.44} &\cellcolor{lightgreen} \textbf{56.23} &\cellcolor{lightgreen} \textbf{90.95} &\cellcolor{lightgreen} \textbf{72.21} \\
\bottomrule
\end{tabular}
\end{adjustbox}
\end{table}

\paragraph{Result on ACEBench.}

Table~\ref{tab:acebench} reports that from an overall perspective on ACEBench, R2IF achieves the highest Overall accuracy across all four backbones.

R2IF's gains are mainly concentrated in Atom and Single-turn, which more directly reflect tool decision quality. On Atom, R2IF clearly surpasses Raw and SFT for every backbone and generally outperforms Binary Reward and ToolRL, suggesting the proposed rewards improve atomic tool decisions. On Single-turn, R2IF also leads on most backbones, with especially strong gains on Qwen2.5-7B and Llama3.2-3B, highlighting better support for end-to-end parameter construction.

Importantly, R2IF maintains strong rejection performance on the Irrelevant subset without trading off core tool-invocation accuracy. Its Overall improvements are therefore not driven by over-optimizing Irrelevant cases, but by more balanced gains on the primary invocation settings (Atom and Single-turn), aligning with ACEBench's goal of assessing comprehensive tool-use capability rather than single-metric optimization.

\FloatBarrier
\begin{table}[t]
\caption{The evaluation results on the ACEBench}
\label{tab:acebench}
\centering
\footnotesize
\setlength{\tabcolsep}{3pt}

\begin{adjustbox}{center,max width=0.75\columnwidth}
\begin{tabular}{llcccc}
\toprule
Model & Method & Atom & Single-turn & Irrelevant & Overall \\
\midrule

\multirow{5}{*}{\shortstack[l]{Qwen2.5-1.5B-Instruct}}
& Raw           & 43.70 & 13.50 & 30.00 & 29.07 \\
& SFT           & 61.30 & 23.00 & 16.00 & 33.43 \\
& Binary Reward & 43.30 & 14.50 & 78.00 & 45.27 \\
& ToolRL        & 52.30 & 26.00 & 88.00 & \uline{55.43} \\
& R2IF \cellcolor{lightgreen}      &\cellcolor{lightgreen} 64.70 &\cellcolor{lightgreen} 20.50 & \cellcolor{lightgreen}88.00 & \textbf{57.73} \cellcolor{lightgreen}\\
\midrule

\multirow{5}{*}{\shortstack[l]{Qwen2.5-3B-Instruct}}
& Raw           & 52.00 & 25.00 & 90.00 & 55.67 \\
& SFT           & 63.70 & 29.50 & 44.00 & 45.73 \\
& Binary Reward & 49.00 & 20.50 & 60.00 & 43.17 \\
& ToolRL        & 58.00 & 31.50 & 96.00 & \uline{61.83} \\
& R2IF \cellcolor{lightgreen}        & 70.70\cellcolor{lightgreen} & 26.00\cellcolor{lightgreen} & 90.00\cellcolor{lightgreen} & \cellcolor{lightgreen}\textbf{62.23} \\
\midrule

\multirow{5}{*}{\shortstack[l]{Qwen2.5-7B-Instruct}}
& Raw           & 57.30 & 40.00 & 88.00 & 61.77 \\
& SFT           & 71.30 & 54.50 & 94.00 & 73.27 \\
& Binary Reward & 76.70 & 49.00 & 96.00 & 73.90 \\
& ToolRL        & 76.30 & 56.50 & 96.00 & \uline{76.27} \\
& R2IF \cellcolor{lightgreen}         &\cellcolor{lightgreen} 78.00 &\cellcolor{lightgreen} 57.50 &\cellcolor{lightgreen} 96.00 & \cellcolor{lightgreen}\textbf{77.17} \\
\midrule

\multirow{5}{*}{\shortstack[l]{Llama3.2-3B-Instruct}}
& Raw           & 45.30 & 17.00 & 14.00 & 25.43 \\
& SFT           & 67.30 & 34.50 & 54.00 & 51.93 \\
& Binary Reward & 72.00 & 47.50 & 90.00 & \uline{69.83} \\
& ToolRL        & 65.70 & 39.00 & 88.00 & 64.23 \\
& R2IF \cellcolor{lightgreen}         &\cellcolor{lightgreen} 73.70 &\cellcolor{lightgreen} 51.00 &\cellcolor{lightgreen} 94.00 &\cellcolor{lightgreen} \textbf{72.90} \\
\bottomrule
\end{tabular}
\end{adjustbox}
\end{table}

\begin{table*}[t]
\caption{Average CoT Effectiveness on BFCL and ACEBench. We test the ACE of the CoT generated during previous evaluations. The table reports the difference in success rates between the student model conditioned on the generated reasoning prefix and the baseline performance, across different models and methods. Positive values indicate that the reasoning improves tool-call performance, while negative values suggest it is not helpful.}
\label{tab:ace}
\centering
\footnotesize
\setlength{\tabcolsep}{3pt}

\begin{adjustbox}{center,max width=0.9\textwidth}
\begin{tabular}{llcccccccc}
\toprule
\multirow{2}{*}{Model} & \multirow{2}{*}{Method}
& \multicolumn{4}{c}{\textbf{BFCL}}
& \multicolumn{4}{c}{\textbf{ACEBench}}\\
\cmidrule(lr){3-6} \cmidrule(lr){7-10}
& 
& Overall& Non-live & Live & Irrelevance 
& Overall & Atom & Single Turn & Irrelevant \\

\midrule
\multirow{5}{*}{Qwen2.5-1.5B-Instruct}
& Raw             & -0.14 & -0.32 & -0.26 & 0.17 & -0.43 & -0.64 & -0.72 & 0.06 \\
& SFT            & \uline{-0.13} & -0.31 & -0.25 & 0.18  & -0.48 & -0.45 & -0.69 & -0.30 \\
& Binary Reward   & -0.15 & -0.31 & -0.22 & 0.09  & -0.40 & -0.69 & -0.71 & 0.21 \\
& ToolRL          & -0.14 & -0.16 & -0.06 & -0.18  & \uline{-0.39} & -0.68 & -0.72 & 0.23 \\
& R2IF \cellcolor{lightgreen}       & \cellcolor{lightgreen}\textbf{-0.03} &\cellcolor{lightgreen} -0.16 &\cellcolor{lightgreen} -0.05 &\cellcolor{lightgreen} 0.13  & \cellcolor{lightgreen}\textbf{-0.34} &\cellcolor{lightgreen} -0.45 & \cellcolor{lightgreen}-0.66 &\cellcolor{lightgreen} 0.10 \\

\midrule

\multirow{5}{*}{Qwen2.5-3B-Instruct}
& Raw             & -0.07 & -0.21 & -0.09 & 0.09 & -0.35 & -0.54 & -0.65 & 0.15 \\
& SFT            & \uline{-0.08} & -0.12 & -0.02 & -0.09   & -0.44 & -0.42 & -0.68 & -0.21 \\
& Binary Reward  & -0.22 & -0.57 & -0.33 & 0.24  & \uline{-0.41} & -0.74 & -0.75 & 0.25 \\

& ToolRL         & -0.23 & -0.60 & -0.32 & 0.22  & \uline{-0.41} & -0.71 & -0.74 & 0.21 \\

& R2IF \cellcolor{lightgreen}          &\cellcolor{lightgreen} \textbf{0.02} &\cellcolor{lightgreen} -0.11 &\cellcolor{lightgreen} -0.03 & \cellcolor{lightgreen}0.18  &\cellcolor{lightgreen} \textbf{-0.29} &\cellcolor{lightgreen} -0.39 &\cellcolor{lightgreen} -0.68 &\cellcolor{lightgreen} 0.21 \\

\midrule
\multirow{5}{*}{Llama3.2-3B-Instruct}
& Raw             & -0.20 & -0.42 & -0.27 & 0.09  & -0.44 & -0.67 & -0.71 & 0.07 \\

& SFT             & \uline{-0.08} & -0.08 & -0.07 & -0.10  & -0.40 & -0.43 & -0.66 & -0.10 \\
& Binary Reward  & -0.12 & -0.30 & -0.21 & 0.14  & \uline{-0.39} & -0.70 & -0.71 & 0.23 \\

& ToolRL         & -0.18 & -0.43 & -0.30 & 0.20  & \uline{-0.39} & -0.70 & -0.73 & 0.25 \\

& R2IF \cellcolor{lightgreen}         & \cellcolor{lightgreen}\textbf{0.05} &\cellcolor{lightgreen} -0.05 &\cellcolor{lightgreen} 0.07 &\cellcolor{lightgreen} 0.14 &\cellcolor{lightgreen} \textbf{-0.28} &\cellcolor{lightgreen} -0.38 &\cellcolor{lightgreen} -0.66 &\cellcolor{lightgreen} 0.21 \\

\bottomrule
\end{tabular}
\end{adjustbox}
\end{table*}

\paragraph{Why the Method with High Accuracy Performance gets low ACE.}
Comparing Table~\ref{tab:bfcl-main}, Table~\ref{tab:acebench}, and Table~\ref{tab:ace}, we observe some methods achieve high accuracy but relatively low ACE scores, such as Qwen2.5-3B with the Binary Reward method.

On BFCL, Qwen2.5-3B's Binary Reward method achieves high accuracy, especially in Non-live and Live settings, but its ACE score remains low, even negative. This suggests that while the model makes correct tool calls, its reasoning is shallow and lacks depth, relying on simple rules rather than thorough analysis.

Low ACE scores indicate inadequate reasoning support for tool-call decisions. While the model may produce correct results, its reasoning lacks transparency, which undermines support for more complex tasks. In contrast, R2IF not only focuses on accuracy but ensures reasoning stability and completeness.

\subsection{Ablation Study}
\label{sec:ablation}
Table~\ref{tab:ablation} presents ablation results showing how training pipeline components jointly enable robust, interpretable function-calling. Across Llama3.2-3B/Qwen2.5-7B and BFCL/ACEBench, removing any single component consistently degrades performance—indicating R2IF's gains stem from complementary, non-redundant signals.

\begin{table*}[t]
\caption{Ablation results on BFCL and ACEBench.
We remove individual components of R2IF, including the SMV reward (wo-smv), the CER reward (wo-cer), and the SFT warm-start (wo-sft). }
\label{tab:ablation}
\centering
\footnotesize
\setlength{\tabcolsep}{3pt}

\begin{adjustbox}{center,max width=0.9\textwidth}
\begin{tabular}{llcccccccc}
\toprule
\multirow{2}{*}{Model} & \multirow{2}{*}{Method}
& \multicolumn{4}{c}{\textbf{BFCL}}
& \multicolumn{4}{c}{\textbf{ACEBench}}\\
\cmidrule(lr){3-6} \cmidrule(lr){7-10}
& 
& Overall& Non-live & Live & Irrelevance 
& Overall & Atom & Single Turn & Irrelevant \\

\midrule
\multirow{5}{*}{Llama3.2-3B-Instruct}
& R2IF   \cellcolor{lightgreen}          &\cellcolor{lightgreen} \textbf{72.21} &\cellcolor{lightgreen} 69.44 &\cellcolor{lightgreen} 56.23 &\cellcolor{lightgreen} 90.95    &\cellcolor{lightgreen} \textbf{72.90} &\cellcolor{lightgreen} 73.70 &\cellcolor{lightgreen} 51.00 &\cellcolor{lightgreen} 94.00  \\
& wo-smv           & \uline{69.26} & 67.69 & 52.54 & 87.55    & \uline{72.40} & 73.70 & 45.50 & 98.00 \\
& wo-cer           & 68.35 & 66.63 & 48.60 & 89.82    & 70.60 & 72.30 & 39.50 & 100.00 \\
& wo-sft           & 66.00 & 65.00 & 45.46 & 87.55    & 67.33 & 65.00 & 45.00 & 92.00  \\
\midrule

\multirow{5}{*}{Qwen2.5-7B-Instruct}
& R2IF  \cellcolor{lightgreen}          &\cellcolor{lightgreen} \textbf{72.14} &\cellcolor{lightgreen} 59.50 &\cellcolor{lightgreen} 63.33 &\cellcolor{lightgreen} 93.59    &\cellcolor{lightgreen} \textbf{77.17} &\cellcolor{lightgreen} 78.00 &\cellcolor{lightgreen} 57.50 &\cellcolor{lightgreen} 96.00  \\
& wo-smv           & \uline{71.99} & 60.19 & 59.85 & 95.94    & \uline{76.90} & 81.70 & 53.00 & 96.00 \\
& wo-cer           & 71.84 & 58.63 & 61.81 & 95.08    & 76.57 & 78.70 & 55.00 & 96.00 \\
& wo-sft           & 68.91 & 57.56 & 55.33 & 93.84    & 74.93 & 76.30 & 52.50 & 96.00  \\

\bottomrule
\end{tabular}
\end{adjustbox}
\end{table*}

\paragraph{Why we need SFT.}
SFT warm-start is more critical for smaller models. As shown in the results, removing SFT causes a larger accuracy drop for the 3B backbone. Compared to 7B models, smaller models possess weaker priors for instruction following and structured outputs, and thus are more prone to early-stage formatting instability, missing fields, or output drift, which reduces the fraction of executable trajectories. Moreover, binary rewards often act as a hard gate: malformed outputs receive little to no learning signal, pushing small models more frequently into a near-zero-reward regime and further impairing exploration efficiency. Therefore, SFT warm-start can establish basic output templates and tool-call syntax, enabling the policy to reliably produce evaluable trajectories; RL can then refine decision quality and reasoning–action alignment on top of this foundation.

\paragraph{Effect of $\mathcal{R}_{\text{smv}}$.}
In contrast, the Specification–Modification–Value (SMV) reward primarily boosts fine-grained parameter decision accuracy. Ablating SMV consistently hurts overall accuracy and BFCL Live performance, reflecting its role in grounding reasoning to concrete parameter specifications and transformations. Moreover, removing SMV slightly increases Irrelevance scores in some cases, suggesting strict parameter-level supervision may marginally trade off with conservative rejection behavior. This highlights SMV targets argument construction precision rather than high-level tool selection.

\paragraph{Effect of $\mathcal{R}_{\text{cer}}$.}
The Chain-of-Thought Effectiveness Reward plays a critical role in stabilizing reasoning quality, particularly in challenging settings. Removing CER leads to a pronounced decline on BFCL Live and ACEBench Single-Turn subsets—known to require stronger generalization and precise reasoning under distribution shift. This indicates CER does not merely encourage verbose reasoning, but actively filters for reasoning prefixes useful for correct tool calls.

\subsection{Further analysis}
\paragraph{Scalability.}

\begin{figure}[t]
  \includegraphics[width=\columnwidth]{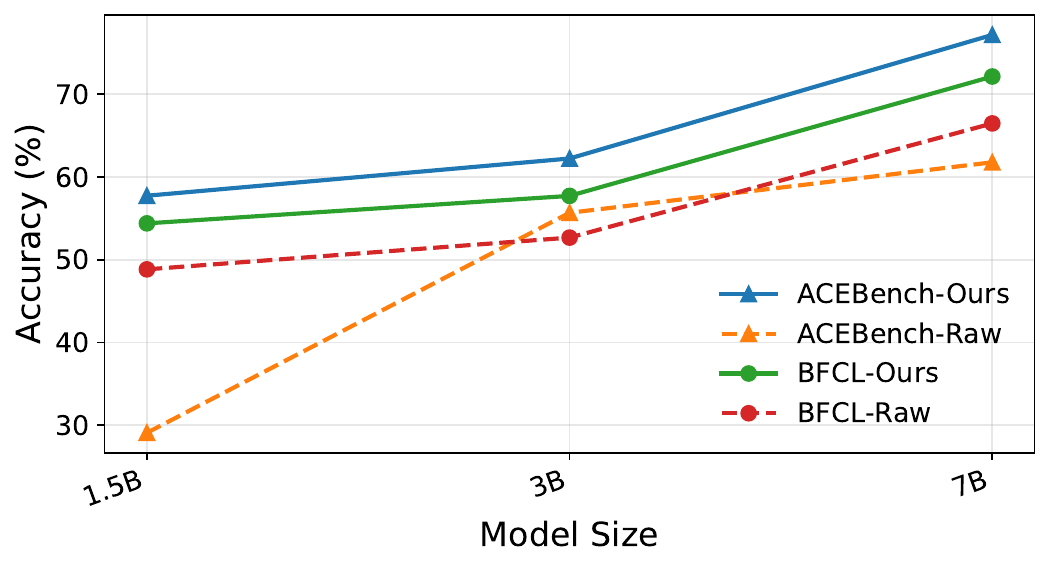}
  \caption{Scaling performance across model sizes using the Qwen2.5-Instruct series.}
  \label{fig:scaling}
\end{figure}
Figure~\ref{fig:scaling} summarizes how performance scales with backbone size on both ACEBench and BFCL. Overall, accuracy increases monotonically from 1.5B to 7B for both the raw instruct models and our trained models, indicating that our approach is compatible with standard scaling trends rather than trading off capacity for alignment. More importantly, our method consistently improves over the raw baselines across all sizes and both benchmarks.

\paragraph{Generalizability.}
Furthermore, our method exhibits strong \emph{generalizability} across model families, not limited to the Qwen2.5 series. We observe consistent and substantial improvements on \textbf{Llama3.2-3B} as well. On BFCL, our approach boosts the Overall score from 37.59\% to \textbf{72.21\%} ($+34.62\%$) while maintaining high Irrelevance performance. On ACEBench, we also improve the Overall accuracy. These results indicate that our reward design can transfer effectively across architectures and evaluation distributions.

\section{Conclusion}

We present R2IF, a reasoning-aware reinforcement learning framework that enhances the interpretability and effectiveness of function calling in LLMs. By aligning reasoning with tool-call decisions, R2IF ensures that predictions are based on valid reasoning. 
Using a composite reward system, we strengthen the link between reasoning and decision-making. Our results demonstrate the value of structured reasoning supervision in improving both tool-call accuracy and reasoning stability, ensuring that outputs are both accurate and interpretable.

\section*{Limitations}
Our framework has several limitations. First, the current evaluation of R2IF is limited to single-turn function query tasks, which restricts our understanding of its performance in multi-turn scenarios. In tasks involving multiple queries or interactions, where reasoning and function calls span across different steps, the model's ability to maintain coherent reasoning and accurate tool calls remains untested. This limitation suggests that the framework's generalizability to more complex, interactive environments is still unclear and requires further exploration.Second, the reward design of R2IF is highly tailored to specific function calling tasks, which may hinder its applicability to tasks that require different forms of reasoning or decision-making. Since the framework's rewards are optimized for particular task constraints and parameter specifications, its performance may vary significantly depending on the type of tool being invoked. As a result, R2IF's scalability and versatility in other LLM applications, where task structures or tools differ, may be limited.

\section*{Acknowledgments}
This research is supported by National Science and Technology Major Project (No.2025ZD1606804), the Ministry of Industry and Information Technology of China, the Key Program of National Natural Science Foundation of China (No. 62432007) and Shanghai Trusted Industry Internet Software Collaborative Innovation Center.

\bibliography{main}

\newpage
\appendix

\section{Multi-turn Interactive Evaluation}
\label{app:multiturn}

To assess whether the proposed reward design generalizes beyond single-turn function calling, we conduct a multi-turn interactive case study on 50 samples, comparing Raw, SFT, Binary-Reward, ToolRL, and R2IF.

\begin{table}[htbp]
\centering
\caption{The evaluation result on the multi-turn interactive case.}
\small
\renewcommand{\arraystretch}{1.1}
\setlength{\tabcolsep}{4pt}
\begin{tabular}{lccccc}
\hline
Model & Raw & SFT & Binary & ToolRL & \textbf{R2IF} \\
\hline
Llama3.2-3B & 4.00 & 8.00 & 2.00 & 4.00 & \textbf{10.00} \\
Qwen2.5-7B  & 8.00 & 10.00 & 14.00 & \textbf{16.00} & \textbf{16.00} \\
\hline
\end{tabular}
\label{tab:multiturn}
\end{table}

Our method remains effective in multi-turn settings: it achieves the best performance on Llama3.2-3B and matches the strongest baseline on Qwen2.5-7B.

Overall, these results suggest that the benefits of R2IF are not limited to single-turn execution and persist when state updates accumulate over an interaction history.

\section{Dependency on Student Evaluator and Sampling Robustness of CER}
\label{app:student}

\subsection{Selection of student model}
\label{app:student-validity}

CER evaluates whether a reasoning prefix increases the downstream success rate of a student model. For this estimator to be meaningful, the student model must satisfy two conditions: (1) it must be able to produce executable tool calls when conditioned on a reasoning prefix, and (2) it must not systematically collapse to rejection behavior on non-irrelevance instances.

We evaluate three candidate student models on 200 samples: Qwen2.5-1.5B, Qwen2.5-3B, and Llama3.2-3B. Table~\ref{tab:student-validity} reports continuation-based success rates under reasoning prefixes.

\begin{table}[htbp]
\centering
\caption{Validity check for student models.}
\small
\renewcommand{\arraystretch}{1.1}
\setlength{\tabcolsep}{4pt}
\begin{tabular}{lcc}
\hline
Student Model & Non-Irrel. & Irrel. \\
\hline
Llama3.2-3B  & \textbf{53.68} & \textbf{75.00} \\
Qwen2.5-1.5B & 0.14  & 92.81 \\
Qwen2.5-3B   & 0.73  & 98.75 \\
\hline
\end{tabular}
\label{tab:student-validity}
\end{table}

The Qwen student models frequently default to the rejection response even when tool invocation is required, causing continuation-based success rates on non-irrelevance cases to degenerate to near-zero constants. Under this behavior, CER becomes largely uninformative. In contrast, Llama3.2-3B preserves meaningful continuation variability while avoiding systematic refusal bias. We therefore adopt it as the student evaluator in our experiments.

\subsection{Sampling Robustness of CER}
\label{app:cer-robustness}

We further examine whether CER is sensitive to sampling hyperparameters. Using the same student model and fixing the common instance set to 200 examples, we vary temperature, top-$p$, and the number of continuations $K$. Let $C_0$ denote the configuration used in the main experiments and other sampling configurations are listed in Table~\ref{tab:cer-config}.

We then compare per-instance CER rankings against $C_0$.

Across configurations, CER exhibits strong ranking stability, limited numerical drift, and no dependence on a narrow temperature or top-$p$ regime. Additional statistics further support this robustness: sign agreement ranges from 0.845 to 0.905, $\Pr(|\Delta| \le 0.2) \approx 0.91\text{-}0.94$, and $\mathbb{E}[|\Delta|] \approx 0.052\text{-}0.076$.

\begin{table}[htbp]
\centering
\caption{Sampling configurations for CER robustness analysis.}
\small
\renewcommand{\arraystretch}{1.1}
\setlength{\tabcolsep}{5pt}
\begin{tabular}{lccc}
\hline
Configuration & Temperature & Top-$p$ & $K$ \\
\hline
$C_0$ & 0.4 & 1.0 & 5 \\
$C_1$ & 0.4 & 0.9 & 5 \\
$C_2$ & 0.7 & 1.0 & 5 \\
$C_3$ & 0.7 & 0.9 & 5 \\
$C_4$ & 0.4 & 1.0 & 10 \\
\hline
\end{tabular}
\label{tab:cer-config}
\end{table}

\begin{table}[htbp]
\centering
\caption{Ranking stability of CER under different sampling configurations.}
\small
\renewcommand{\arraystretch}{1.1}
\setlength{\tabcolsep}{4pt}
\begin{tabular}{lcc}
\hline
Comparison & Spearman $\rho$ & Kendall $\tau$ \\
\hline
$C_0$ vs $C_1$ & 0.944 & 0.897 \\
$C_0$ vs $C_2$ & 0.915 & 0.854 \\
$C_0$ vs $C_3$ & 0.896 & 0.833 \\
$C_0$ vs $C_4$ & 0.936 & 0.884 \\
\hline
\end{tabular}
\label{tab:cer-rank}
\end{table}

\section{Incremental Contribution Beyond ToolRL}
\label{app:beyond-toolrl}

To isolate the contribution of CER and SMV beyond ToolRL-style outcome rewards, we replace the binary reward component in R2IF with ToolRL's correctness-plus-format reward, denoted as R2IF-toolrl.

\begin{table}[htbp]
\centering
\caption{Incremental contribution of CER and SMV beyond ToolRL-style reward supervision.}
\label{tab:beyond-toolrl}
\small
\renewcommand{\arraystretch}{1.1}
\setlength{\tabcolsep}{4pt}
\begin{tabular}{clcc}
\hline
Model & Method & BFCL & ACEBench \\
\hline
\multirow{3}{*}{Llama3.2-3B}
& ToolRL       & 66.26 & 64.23 \\
& R2IF-toolrl & 66.73 & 72.50 \\
& R2IF         & \textbf{72.21} & \textbf{72.90} \\
\hline
\multirow{3}{*}{Qwen2.5-3B}
& ToolRL       & 51.44 & 61.83 \\
& R2IF-toolrl & 57.14 & 61.73 \\
& R2IF         & \textbf{57.71} & \textbf{62.23} \\
\hline
\end{tabular}
\end{table}

As shown in Table~\ref{tab:beyond-toolrl}, adding CER and SMV on top of ToolRL-style reward supervision yields consistent gains across backbones, which indicate that CER and SMV provide complementary process-level credit assignment beyond ToolRL-style outcome rewards.

\section{Computational Overhead Analysis}
\label{app:compute-overhead}

We analyze the computational overhead of our method from two perspectives: training cost and inference cost.

\subsection{Training Overhead}
\label{app:training-overhead}

We first measure the training overhead introduced by CER over 9 epochs and 36 steps. Table~\ref{tab:cer-overhead} reports the average training time per step.

\begin{table}[htbp]
\centering
\caption{Average training time per step (seconds).}
\small
\renewcommand{\arraystretch}{1.1}
\setlength{\tabcolsep}{4pt}
\begin{tabular}{lcc}
\hline
Model & With CER & Without CER \\
\hline
Llama3.2-3B & 251.29 & 225.52 \\
Qwen2.5-7B  & 612.64 & 595.19 \\
\hline
\end{tabular}
\label{tab:cer-overhead}
\end{table}

Enabling CER increases per-step training time by 11.4\% on Llama3.2-3B and 2.9\% on Qwen2.5-7B. These results indicate that the computational cost is driven mainly by model scale, whereas CER introduces only a limited additional overhead.

\subsection{Inference Overhead}
\label{app:inference-overhead}

We next examine the inference overhead by reporting average reasoning length on the BFCL test set (3,475 cases). Table~\ref{tab:reason_length} reports the average number of reasoning tokens across methods and backbones.

\begin{table}[htbp]
\centering
\caption{Average reasoning length (tokens) on BFCL.}
\small
\renewcommand{\arraystretch}{1.1}
\setlength{\tabcolsep}{3.5pt}
\resizebox{\columnwidth}{!}{
\begin{tabular}{lccccc}
\hline
Model & Raw & SFT & Binary & ToolRL & R2IF \\
\hline
Llama3.2-3B & 118.54 & 178.34 & 98.86 & 91.89 & 163.00 \\
Qwen2.5-3B  & 245.33 & 376.92 & 142.83 & 147.37 & 278.03 \\
\hline
\end{tabular}
}
\label{tab:reason_length}
\end{table}

The results in Table~\ref{tab:reason_length} suggest two points. First, the absolute reasoning length of R2IF remains moderate and stays within the same scale as SFT reasoning. Second, the increase relative to ToolRL is bounded: +71 tokens for Llama3.2-3B and +131 tokens for Qwen2.5-3B. This increase is expected, since the objective explicitly encourages exposing the chain
\[
\text{Specification} \rightarrow \text{Modification} \rightarrow \text{Value},
\]
which requires additional tokens to make otherwise implicit transformations explicit.

Overall, the additional inference cost remains controlled, while yielding more interpretable and decision-grounded reasoning for function calling.

\section{Human Evaluation of Reasoning Interpretability}
\label{app:humaneval}
To complement automatic evaluation, we conduct a human study of reasoning interpretability with the framework\cite{bhambri2025cognitivelyinterpretablereasoningtraces}. We evaluate four dimensions: \emph{Predictability}, \emph{Comprehensibility}, \emph{Interpretability}, and \emph{Faithfulness}.

The study uses the Llama3.2-3B backbone and compares Raw, Binary-Reward, ToolRL, and R2IF. We recruited 8 evaluators, each of whom assessed 5 tasks under the Llama3.2-3B setting.

\begin{table}[htbp]
\centering
\caption{Human evaluation results on reasoning interpretability.}
\small
\renewcommand{\arraystretch}{1.1}
\setlength{\tabcolsep}{3.5pt}
\resizebox{\columnwidth}{!}{
\begin{tabular}{lcccc}
\hline
Dimension & R2IF & ToolRL & Binary & Raw \\
\hline
Predictability    & \textbf{4.850} & 2.875 & 2.450 & 2.825 \\
Comprehensibility & \textbf{4.825} & 3.300 & 2.725 & 3.150 \\
Interpretability  & \textbf{4.800} & 2.900 & 2.575 & 2.875 \\
Faithfulness      & \textbf{4.775} & 2.650 & 2.275 & 2.775 \\
\hline
\end{tabular}
}
\label{tab:human-eval}
\end{table}

As illustrated in Table~\ref{tab:human-eval}, in all four dimensions, our method outperforms the baselines by a large margin. All pairwise comparisons between R2IF and the other methods are statistically significant ($p < 0.05$). In addition, the evaluation exhibits strong reliability, with Cronbach's $\alpha > 0.7$ for all dimensions. These results provide direct evidence that our method improves the interpretability of reasoning traces.

\section{Experiment}

\subsection{SFT Setting}
We conduct supervised fine-tuning (SFT) using the TRL framework, and summarize the hyperparameter configuration in Table~\ref{tab:sft_config}.

\FloatBarrier
\begin{table}[htbp]
\centering
\small
\caption{Configuration for SFT training}
\label{tab:sft_config}
\begin{tabular}{lc}
\toprule
\textbf{Hyperparameter} & \textbf{Value} \\
\midrule
Train Batch Size        & 512 \\
Max Prompt Length       & 3072 \\
Max Response Length     & 1024 \\
Learning Rate           & \num{1e-5} \\
Total Epochs            & 2 \\
\bottomrule
\end{tabular}
\end{table}

\subsection{GRPO Setting}
The training was conducted using the Verl framework, and the configuration for GRPO training is summarized in Table~\ref{tab:grpo_config}. Below are the hyperparameters used for training:

\FloatBarrier
\begin{table}[htbp]
\centering
\small
\caption{Configuration for GRPO training}
\label{tab:grpo_config}
\begin{tabular}{cc}
\toprule
\textbf{Hyperparameter} & \textbf{Value} \\
\midrule
Train Batch Size & 512 \\
Validation Batch Size & 128 \\
Max Prompt Length & 3072 \\
Max Response Length & 1024 \\
Learning Rate & \num{1e-6} \\
PPO Mini Batch Size & 128 \\
GPU Memory Utilization & 0.6 \\
Number of Rollouts & 5 \\
Total Epochs & 9 \\
\bottomrule
\end{tabular}
\end{table}

\section{Prompt}
\subsection{Prompt for LLM Inference}
The System Prompt in Figure~\ref{fig:prompt4inference}, with its predefined instructions and constraints, standardizes the model's behavior and output. It ensures consistent learning during training and uniform evaluation based on a unified set of rules.

\begin{figure*}[htbp]
    \begin{promptbox}{Prompt for LLM Inference.}
You are an expert in composing functions. You are given a question and a set of possible functions. \\
Based on the question, you will need to make one or more function/tool calls to achieve the purpose.

\vspace{0.5em}
In each action step, you MUST: \\
1. Think about the 2-step reasoning process in the mind and enclosed your reasoning within <reason></reason> tags. \\
\hspace*{1.5em} - Step1: Analyze the question and highlight the chosen function in the format of: \#fun\_name\# \\
\hspace*{1.5em} - Step2: For each function you decide to call, you should analyze the value of each parameter in the format of: - params\_name1: analysis\_process \\
2. Then, provide the final function call with function names and arguments in the format of: <tool>[\{"name": "func\_name1", "arguments": \{"params\_name1": params\_value1, "params\_name2": params\_value2...\}\}, \{"name": "func\_name2", "arguments": \{\}\}]</tool> \\
3. Make sure both the reasoning and the tool call steps are included together in one single reply. 

\vspace{0.5em}
Output format: \\
<reason> \\
Step1: ... \\
Step2: \\
For \#fun\_name1\#: \\
- params\_name1: analysis\_process \\
... \\
</reason> \\
<tool>[\{"name": "func\_name1", "arguments": \{"params\_name1": params\_value1, "params\_name2": params\_value2...\}\}, \{"name": "func\_name2", "arguments": \{\}\}]</tool> \\
You SHOULD NOT include any other text in the response.

\vspace{0.5em}
Important Notes \\
1. If the given question lacks the arguments required by the function, point it out. If this required parameter has enum option, you can choose a single unambiguous match from listed option. \\
2. If none of the function can be used, write it in <tool>None of function can be used</tool>.

    \end{promptbox}
    \caption{Prompt for LLM Inference.}
    \label{fig:prompt4inference}
\end{figure*}

\subsection{Prompt for Construction of Ground-Truth Document}
\label{prompt4smv}
Under the prompt in Figure~\ref{fig:prompt4smv}, a large language model is used to generate SMV annotations as the ground-truth document. Given the user query, tool documentation, and ground-truth tool-call result as input, the model produces parameter-level SMV annotations, including the relevant specification, necessary modification and final value for each argument.

\begin{figure*}[htbp]
    \begin{promptbox}{Prompt for Construction of Ground-Truth Document.}

You are an expert in composing functions. You are given a question, a set of functions doc and a right reference of tool call. \\
You should explore the specification, modification and more argument value may exist for each toolcall in order. \\
And give me the reason why these specifications and modifications exist. no more than three sentences. \\
The specification are some special format requirements in the parameter description \\
\hspace*{1.5em} - write \texttt{no spec} if here is no specification or not mentioned in the reason \\
\hspace*{1.5em} - be careful with the e.g. tag that some parameters' description with it do not mean they have format specification \\
\hspace*{1.5em} - no more than 15 words \\
\hspace*{1.5em} - all the specification must be copied from description without any change

\vspace{0.5em}
The modification is the action that convert the origin value (text from question) in the final state \\
\hspace*{1.5em} - include change process from the \texttt{original value} to \texttt{modified value} \\
\hspace*{1.5em} - write \texttt{no modify} if change nothing \\
\hspace*{1.5em} - no more than 15 words

\vspace{0.5em}
If there is a modification, then there must be a specification. If there is no modification, it doesn't necessarily mean there is no specification.

\vspace{0.5em}
Note that the provided function is in Python 3 syntax.

\vspace{0.5em}
The specification includes: \\
\hspace*{1.5em} 1 \textbf{units}: specify expected unit systems (e.g., SI units), allowable representations such as symbols vs. full names, and whether spacing is required between value and unit (e.g., "10 kg" vs "10kg"). \\
\hspace*{1.5em} 2 \textbf{dates/times}: only choose one from following (e.g., YYYY-MM-DD, YYYY/MM/DD, DD/MM/YYYY, or "January 1, 2025"). \\
\hspace*{1.5em} 3 \textbf{percentages}: specify whether values should be expressed as the percentage sign (e.g., "0.5" vs "50" vs "50\%"). \\
\hspace*{1.5em} 4 \textbf{place names / locations}: specify expected string patterns (e.g., "City, State" such as "San Francisco, CA" where applicable, or "City, Country" when no state exists), and indicate whether abbreviated forms of administrative regions should be used. \\
\hspace*{1.5em} 5 \textbf{precision}: specify how precision should be expressed (e.g., number of decimal places, significant digits). \\
......\\

\vspace{0.5em}
Output in the json format: \\
{}[ \\  
\hspace*{1.5em} \{ \\
\hspace*{1.5em} \hspace*{1.5em} "reason": "...", \\
\hspace*{1.5em} \}, \\
\hspace*{1.5em} \{ \\
\hspace*{3em} "name": "func\_name1", \\
\hspace*{3em} "arguments": \{ \\
\hspace*{4.5em} "param\_name1":\{ \\
\hspace*{6em} "specification": spec, \\
\hspace*{6em} "modification": mod, \\
\hspace*{4.5em} \}, \\
\hspace*{4.5em} "param\_name2":\{ \\
\hspace*{6em} .... \\
\hspace*{4.5em} \} \\
\hspace*{3em} \} \\
\hspace*{1.5em} \}, \\
\hspace*{1.5em} \{ \\
\hspace*{3em} "name": "func\_name2", \\
\hspace*{3em} "arguments": \{ \\
\hspace*{4.5em} ... \\
\hspace*{3em} \} \\
\hspace*{1.5em} \} \\
\hspace*{1.5em} ... \\
]

    \end{promptbox}
    \caption{Prompt for Construction of Ground-Truth Document.}
    \label{fig:prompt4smv}
\end{figure*}

\end{document}